# Short Term Load Forecasting Using Deep Neural Networks

Faisal Mohammad
Department of Computer Engineering
Chonbuk National University
Jeonju 561-756, South Korea
mfaisal@jbnu.ac.kr

Ki-Boem Lee
Department of Computer Engineering
Chonbuk National University
Jeonju 561-756, South Korea
keywii@jbnu.ac.kr

Young-Chon Kim
Smart Grid Research Center
Chonbuk National University
Jeonju 561-756, South Korea
yckim@jbnu.ac.kr

*Abstract*— Electricity load forecasting plays an important role in the energy planning such as generation and distribution. However, the nonlinearity and dynamic uncertainties in the smart grid environment are the main obstacles in forecasting accuracy. Deep Neural Network (DNN) is a set of intelligent computational algorithms that provide a comprehensive solution for modelling a complicated nonlinear relationship between the input and output through multiple hidden layers. In this paper, we propose DNN based electricity load forecasting system to manage the energy consumption in an efficient manner. We investigate the applicability of two deep neural network architectures Deep Feed-forward Neural Network (Deep-FNN) and Deep Recurrent Neural Network (Deep-RNN) to the New York Independent System Operator (NYISO) electricity load forecasting task. We test our algorithm with various activation functions such as Sigmoid, Hyperbolic Tangent (tanh) and Rectifier Linear Unit (ReLU). The performance measurement of two network architectures is compared in terms of Mean Absolute Percentage Error (MAPE) metric.

## I. INTRODUCTION

In smart grid, the importance of demand side energy management including aggregated load forecasting is becoming critical. Electrical load forecasting is still a challenging problem due to the complex and nonlinear nature. Furthermore, with the exponential advent of electric vehicles (PEVs, PHEVs etc.) will increase the electric consumption from the demand side largely by 2030 [1-3]. Load forecasting has great impacts on the smart grid applications including energy purchasing, electric vehicle scheduling and planning. Although, with the recent development of internet of things (IoT) and advanced smart meter technology, the ability to record relevant information on a large scale has been done. Traditional methods strive in analyzing such complicated relationships for their limited abilities in supervising nonlinear data [4]. Therefore, short-term load forecasting plays an important role because of its wide applicability to demand side management, energy storage operation, peak load reduction, etc. [5, 6]. Among these, short-term power load forecasting is challenging task due to little time duration and accurate decision. Inaccurate load forecasting harms the scheduling and planning of power systems, which can cause the shortage of energy in market and eventually increases the cost to produce more energy [7].

In the published literatures, power load forecasting methods have been widely studied. A review of artificial neural network for short term load forecasting is presented in [8]. Authors in [9] provide a survey for forecasting models in short-term power load forecasting as shown in fig. 1. In present, deep learning (DL) methods can be found in the literature for performing forecasting tasks. A. Almalaq et al. provides an overview of the DL algorithms and methods that are used in smart grid load forecasting. Deep Belief Networks are used in the work of [10] along with RBM. In [6] the authors shed some light on the study of DNN for STLF by proposing two state-of-art architectures and concluded with a need for large dataset to explore a better DNN structure. The disadvantage of this method is it is very slow to train and require a lot of power. Authors in [11] trained DNN with various combination of activation functions for STLF but have small number of samples for training process which can cause the model to overfit. To avoid overfitting authors in [5] used 5 year ISO-New England electricity load and weather data with time-frequency feature selection to train the proposed DNN models using single activation function. A review for using deep learning in forecasting the load is presented in [12]. [13] developed ANN models for hour ahead load and price forecasts using the load data from ISO New York) that we also use in this paper.

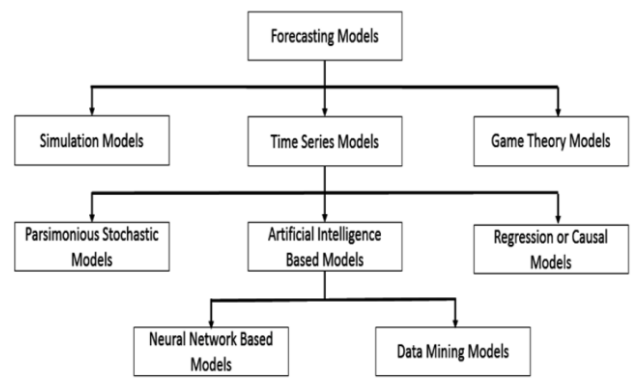

Fig. 1 Forecasting Models [9]

In this work, we train two different state-of art neural network architectures Deep Feed-forward Neural Networks (Deep-FNN) and Deep Recurrent Neural Networks (Deep-RNN) with multiple activation functions like sigmoid, tanh and ReLU, to predict short term electricity load. Achieving higher accuracy in forecasts requires to include all the related factors that affect the overall electricity consumption.

The remaining sections are organized as follows, section II is dedicated for the Deep-FNN and Deep-RNN based models. In section III, the description and pre-processing of the dataset is given, whereas section IV presents the methodology that covers data preprocessing as well as the aspects of data training and prediction models. Section V exhibits the analysis and results obtained from our experimentation. Finally, work is concluded in section VI.

## II. DEEP NEURAL NETWORK ARCHITECTURES

Deep learning is usually implemented using a neural network architecture. The term "deep" refers to the number of layers in the network—the more layers, the deeper the. A deep neural network combines multiple nonlinear processing layers, using simple elements operating in parallel and inspired by biological nervous systems. It consists of an input layer, several hidden layers, and an output layer network shown in fig. 2. The layers are interconnected via nodes, or neurons, with each hidden layer using the output of the previous layer as its input .

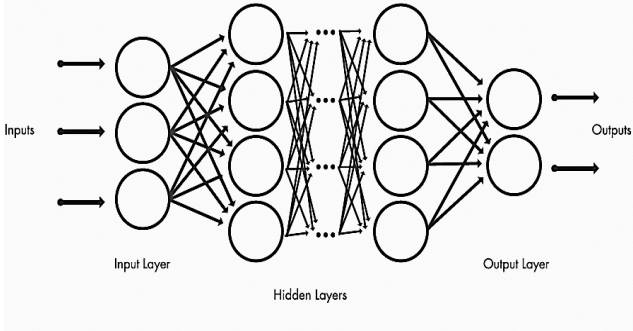

Fig. 2 Deep Neural Networks

### A. Deep Feed Forward Neural Network (Deep-FNN)

Deep feedforward networks also often called feedforward neural networks or multilayer perceptrons (MLPs), are the quintessential deep learning models fig. 3. The goal of a feedforward network is to approximate some function $f$. which maps an input $x$ to $y$. It learns the value of the parameters $\theta$ that result in the best function approximation.

$$y_j = f_j(net_j) \tag{1}$$

$$net_j = \sum_{i=1}^{n} w_{j,i} y_{j,i} + b \tag{2}$$

$$Y = F\left(y_j * w_{j,i}\right) \tag{3}$$

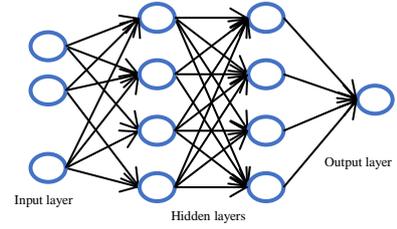

Fig. 3. Deep Feed Forward Neural Network

The output of the neuron is calculated using (1) where $f_j$ is the activation function of neuron $j$ and the net value $net_j$ is the sum of weighted input nodes of that neuron given by (2) where, $y_{j,i}$ is the $i^{th}$ input node of neuron $j$, weighted by $w_{j,i}$, $b$ is the bias weight and $n$ is the number of inputs to the neuron $j$. $f$ is hidden layer function, and $F$ is the output layer transfer function which is considered linear in case of regression. The network learns by adjusting weights to minimize loss function with the large set of training data. The loss function is given by

$$Loss(q|W,B) \tag{4}$$

where q represents each training sample in the dataset and W is the weight matrix and B is the vector of biases.

### B. Deep Recurrent Neural Networks (Deep-RNN)

When feedforward neural networks are extended to include feedback connections, they are called recurrent neural networks. At each time-step of sending input through a recurrent network, nodes receiving input along recurrent edges receive input activations from the current input vector and from the hidden nodes in the network's previous state. The output is computed from the hidden state at the given time-step. The previous input vector at the previous time step can influence the current output at the current time-step through the recurrent connections.

$$y_j^t = f_j(net_j^t) \tag{5}$$

$$net_j = \sum_{i=1}^{n} w_{j,i} y_{j,i}^t + y_j^{t-1} * w_{j,d} \tag{6}$$

$$Y^t = F\left(y_j^t * w_{j,i}\right) \tag{7}$$

Where $y_{j,i}^t$ is $i^{th}$ input node of neuron $j$ at time t, $y_j^t$ is output, $y_j^{t-1}$ is input at time t, $w_{j,i}$ is weight for input layer, $w_{j,d}$ represents weight for time delay input and Wo is the weight for output layer, $f_j$ is hidden layer transfer function and $F$ is the output layer transfer function which is considered linear for regression. The time delay unit is required to hold the output and feedback at the next time step.

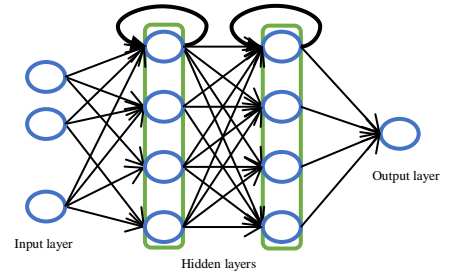

Fig. 4 Deep Recurrent Neural Network

## III. DATA DESCRIPTION AND PRE-PROCESSING

### A. Input and output variables

The electricity load data from 2007 to 2014 is collected from New York Independent System Operator (NYISO). The dataset contains total of 62104 records represents electricity load data for 11 regions of New York state. The weather data is taken from National Climatic Data Center.

Next, aggregated time series load and weather data go through data preprocessing step separately. Data preprocessing contains data cleansing, smoothing, and data restructuring shown in fig. 5. Now the two data sources need to be merged together along a common axis.

Finally, in pre-processing step data restructuring or feature extraction is done to extract the features from the original dataset. Empirical observations done in [3] led to consider factors such as weather, time, holidays, lagged load and load distribution in different time periods to be the most influential to electricity consumption daily.

*1) Temperature:* Both dry bulb and dew point temperature features has been selected to represent temperature effect on the electricity consumption.

*2) Working and Non-Working Days:* In weekdays, electricity consumption is high while on Saturday and Sunday electricity consumption is low and similarly on other social holidays. Based on these observations the "Working Day feature" is selected to draw this effect.

*3) Time Effect:* The electricity consumption is highly dependent over time. Electricity consumption values represent ascending and descending effect before and after the midday respectively. Two features are derived to the express the time dependency as Hour and DayOfWeek.

*4) Lagged Load Effect:* Lagged hourly peak load values contain vital information about specific patterns of users' electricity consumption. This information can be extracted using multiple lagged load values with the features prior day same hour load *L(h, d-1, w)*, prior 1 hour load *L(h-1, d, w)* and prior week same hour load *L(h, d, w-1)*.

### B. Classify training, test and validation datasets

The whole dataset after pre-processing has 10 columns and 62104 rows. This dataset is now divided into training and testing dataset, with 6 years (2007-2013) of data for training and one year data as test set. From the test dataset we select 24 and 168 records for day and week ahead forecasting respectively.

## IV. SYSTEM MODEL

In this paper, we adopt two different DNN models Deep-FNN and Deep-RNN to learn nonlinear relations between weather variables, date, and previous electricity consumptions. Visual description of our proposed system architecture of DNN is shown in fig. 5.

### A. Activation function

The activation function of neural network acts as deciding parameter or transfer function, to transfer weighted inputs to generate the network outputs. Several activation functions are available such as; linear function, step function, logistic function, tangent-hyperbolic (tanh), sigmoid function and rectified linear unit (ReLU) etc. But there is no rule of thumb defined for activation function selection to produce the better result. It can be observed that the change in activation function may affect the output of network. In this work, we implemented our models using three nonlinear activation functions sigmoid, tanh and ReLU.

*1) Sigmoid function*: The sigmoid activation function is very similar to the step function Fig. 6, which acts like threshold. It is a function of form

$$f(x) = \frac{1}{1 + e^{-x}} \qquad (8)$$

where x is input to a neuron. Sigmoid function saturate and kill gradients. It has slow convergence.

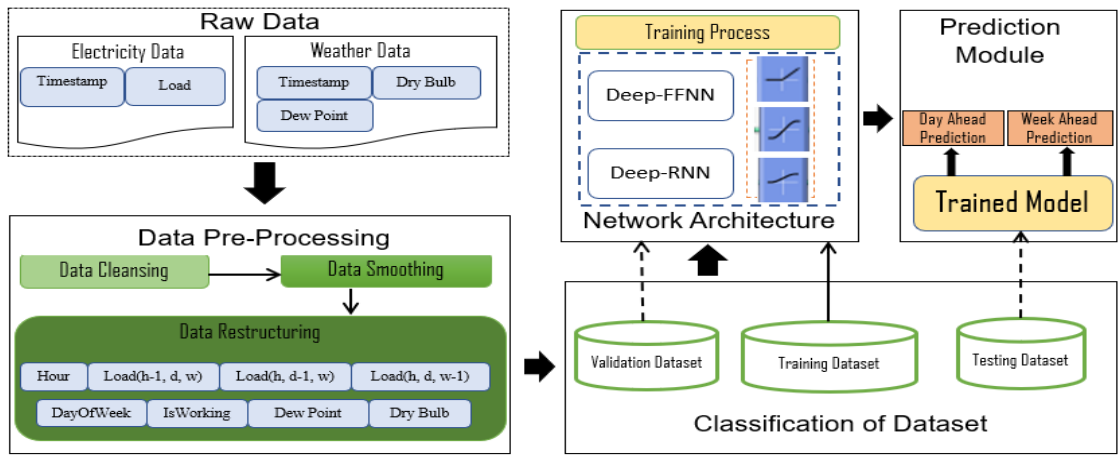

Fig. 5. System Architecture of Load Prediction Model

*2) Hyperbolic Tangent function:* It's mathematical formula is

$$f(x) = \frac{1 - e^{-2x}}{1 + e^{-2x}} \quad (9)$$

Its output is zero centered because the range lies between -1 to 1 i.e. $-1 < f(x) < 1$, fig. 7. Hence optimization is easier in this method hence in practice it is always preferred over Sigmoid function.

*3) Rectified linear unit:* ReLU is similar to the linear function but having zero output across half domain. Recently, it is the widest used activation function in DNN. It avoids and rectifies vanishing gradient problem. Mathematically it is given as

$$f(x) = \max(0, x) \quad (10)$$

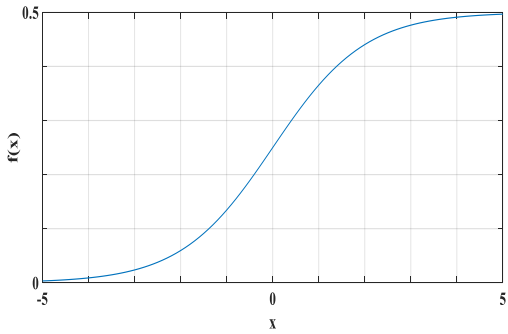
Fig. 6. Sigmoid function

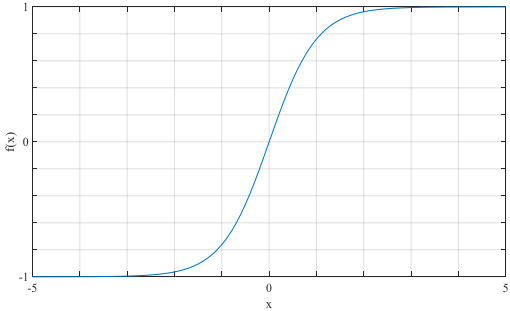
Fig. 7. Hyperbolic tangent function

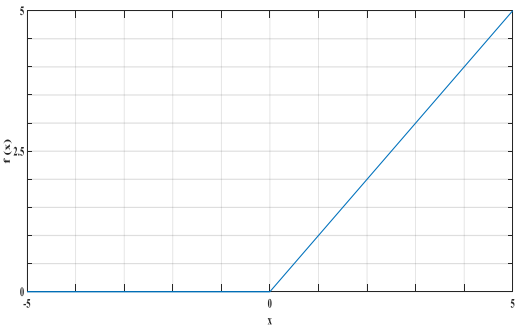
Fig. 8. Rectified linear unit (ReLU)

### B. Network Architecture

The architecture of neural networks is mainly determined by the number of layers and neurons in each layer. The numbers of input neurons and output neurons are automatically fixed by the feature dimension of training data and forecasting period respectively. However, in the case of DNNs, it is not feasible to test all possible combinations of layers and neurons since we need to consider both width and depth of network. We selected 8 neurons in input layer, 10 neurons in each hidden layer for generalization purpose and output layer has one neuron which gives hourly forecast. Hidden layers have activation functions like sigmoid, tanh or ReLU described previously while as the output layer has linear activation funtion to give the predicted output [15].

### C. Optimization algorithm

The Levenberg-Marquardt backpropagation learning algorithm used the gradient descent method to update the weights and biases [16]. The update rule of Levenberg–Marquardt algorithm is given by (10)

$$w_{k+1} = w_k - (J_k^T J_k + \mu I)^{-1} J_k e_k \quad (10)$$

Where: $\mu$ is always positive, called combination coefficient; $I$ is the identity matrix; $J$ is the Jacobian matrix; $k$ is the index of iterations.; $e$ is the error produced in $k^{th}$ iteration.

### D. Performance measurements

The measure of performance used in this work is the mean absolute percent error (MAPE) defined in (10). This error is used to select the best network, which is the one that will be used for forecasting.

$$MAPE = \frac{1}{n} \sum_{i=1}^{n} \frac{L_i^{actual} - L_i^{forecast}}{L_i^{actual}} \times 100\% \quad (11)$$

Where $n$ is the number of samples; $L_i^{actual}$ is $i^{th}$ the actual electricity load; $L_i^{forecast}$ is the $i^{th}$ predicted load.

### V. ANALYSIS AND RESULTS

In this paper, the system model is implemented using Neural Network Toolbox™ in Matlab R2017a. Neural Network Toolbox™ provides algorithms, and simulate both shallow and deep neural networks. Next concern is speed to execute the training process on large datasets, we distribute computations and data across multicore processors on the desktop with Parallel Computing Toolbox™. Several short term load forecasting cases using various combinations of activation functions with Deep-FNN and Deep-RNN investigated areas under

1) *Case 1.* Deep-RNN with sigmoid
2) *Case 2.* Deep-RNN with tanh
3) *Case 3.* Deep-RNN with ReLU
4) *Case 4.* Deep-FNN with sigmoid

5) *Case 5.* Deep-FNN with tanh
6) *Case6.* Deep-FNN with ReLU

Table I shows the data characteristics and other parameters used in our simulation. We forecast load in winter, spring, summer and autumn separately. The proposed models are able to get least MAPE, error in winter season while highest errors are found in summer season as depicted in Table II. This is due to the unexpected variations in electricity consumption because of high temperature and othyer social events in summer season. From table II, it is clear that Deep-RNN with tanh activation function outperformed the other cases .

Table I SIMULATION PARAMETERS

| *Input parameters* | *Values* |
|---|---|
| Total number of samples | 62104 |
| Training data | 53321 |
| Test data | 8783 |
| Maximum number of training epochs | 10000 |
| Training algorithm | trainlm |
| Day of Week | [1 ~ 7] |
| IsWorking | [0 , 1] |
| Hour of Day | [0 ~ 23] |
| Dew Point °F | [-16 ~ 76] |
| Prior day load L (w, d-1, h) MW | [2854 ~ 11447] |
| Prior hour load L (w, d, h-1) MW | [2854 ~ 11447] |
| Prior week load L (w-1, d, h) MW | [2854 ~ 11447] |
| Dry Bulb °F | [4~103] |

Table II SUMMARY OF THE PREDICTION ERROR WITH DEEP-FNN AND DEEP-RNN

| | *Forecast Type* | *MAPE* | | | | | |
|---|---|---|---|---|---|---|---|
| | | *Case 1* | *Case 2* | *Case 3* | *Case 4* | *Case 5* | *Case 6* |
| Winter | day | 0.68 | **0.61** | 0.72 | **0.61** | 0.64 | 0.84 |
| | week | 0.70 | **0.68** | 0.75 | 0.78 | 0.68 | 0.73 |
| Spring | day | 0.73 | **0.65** | 0.80 | 0.75 | 0.74 | 0.82 |
| | week | **0.71** | **0.71** | 0.73 | 0.75 | 0.73 | 0.73 |
| Summer | day | **0.78** | 0.91 | 0.85 | 0.90 | 0.91 | 0.89 |
| | week | 0.85 | **0.73** | 0.78 | 0.85 | 0.83 | 0.80 |
| Autumn | day | 0.80 | 0.71 | 0.67 | 0.71 | **0.66** | 0.77 |
| | week | 0.65 | **0.58** | 0.60 | 0.60 | 0.59 | 0.62 |

## VI. CONCLUSION

This paper investigated the two deep neural networks for short term electricity load forecasting using Levenberg-Marquardt backpropagation algorithm. The original dataset undergoes pre-processing phase to deduce the new features which would be more influential for electricity consumption. Next, we trained the two DNN architectures, Deep-FNN and Deep-RNN using pre-processed dataset to predict the day ahead and week ahead loads. Various activation functions were employed in the hidden layers to train the two DNN architectures. Then, we tested the trained model with the test dataset for 24 and 168 records for all the four seasons. The results indicated that the case 2 viz., Deep-RNN with tanh activation function performs better than other cases in terms of MAPE metric values.

ACKNOWLEDGMENT

This work was supported by the National Research Foundation of Korea (NRF) funded by the Ministry of Science, ICT and Future Planning (2017-004868).